\documentclass[conference]{IEEEtran}
\IEEEoverridecommandlockouts
\usepackage{cite}
\usepackage{amsmath,amssymb,amsfonts}
\usepackage{graphicx}
\usepackage{textcomp}
\usepackage{xcolor}
\usepackage{booktabs}
\usepackage{algorithm}
\usepackage{algpseudocode}
\usepackage{amsmath}
\usepackage{multirow} 
\usepackage{threeparttable}
\usepackage{float}
\usepackage{subfig}
\usepackage{circledsteps}
\def\BibTeX{{\rm B\kern-.05em{\sc i\kern-.025em b}\kern-.08em
    T\kern-.1667em\lower.7ex\hbox{E}\kern-.125emX}}
\begin{document}

\title{Diffusion Policies with Offline and Inverse Reinforcement Learning for Promoting Physical Activity in Older Adults Using Wearable Sensors\\
\thanks{Accepted at ICMLA 2025.}
\thanks{This work was supported in part by the U.S. National Institutes of Health under grants R01MD018025 and 3R01MD018025-02S1, the U.S. National Science Foundation under grants 2426340, 2416727, 2421865, and 2421803.}
}

\author{
\IEEEauthorblockN{
Chang Liu\textsuperscript{1}, 
Ladda Thiamwong\textsuperscript{2}, 
Yanjie Fu\textsuperscript{3}, 
Rui Xie\textsuperscript{1}\textsuperscript{2}\textsuperscript{\dag}
}
\IEEEauthorblockA{\textsuperscript{1}Department of Statistics and Data Science, University of Central Florida, Orlando, FL, USA \\
\textsuperscript{2}College of Nursing, University of Central Florida, Orlando, FL, USA \\
\textsuperscript{3}School of Computing and Augmented Intelligence, Arizona State University, Tempe, AZ, USA \\
Email: chang.liu@ucf.edu, ladda.thiamwong@ucf.edu, yanjie.fu@asu.edu, rui.xie@ucf.edu}
\thanks{\textsuperscript{\dag}Corresponding Author}
}
\maketitle
\IEEEpeerreviewmaketitle
\begin{abstract}
Utilizing offline reinforcement learning (RL) with real-world clinical data is getting increasing attention in AI for healthcare. However, implementation poses significant challenges. Defining direct rewards is difficult, and inverse RL (IRL) struggles to infer accurate reward functions from expert behavior in complex environments. Offline RL also encounters challenges in aligning learned policies with observed human behavior in healthcare applications.
To address challenges in applying offline RL to physical activity promotion for older adults at high risk of falls, based on wearable sensor activity monitoring, we introduce \textbf{K}olmogorov-\textbf{A}rnold \textbf{N}etworks and \textbf{D}iffusion Policies for Offline \textbf{I}nverse Reinforcement Learning (KANDI). By leveraging the flexible function approximation in Kolmogorov-Arnold Networks, we estimate reward functions by learning free-living environment behavior from low-fall-risk older adults (experts), while diffusion-based policies within an Actor-Critic framework provide a generative approach for action refinement and efficiency in offline RL. 
We evaluate KANDI using wearable activity monitoring data in a two-arm clinical trial from our Physio-feedback Exercise Program (PEER) study, emphasizing its practical application in a fall-risk intervention program to promote physical activity among older adults. 
Additionally, KANDI outperforms state-of-the-art methods on the D4RL benchmark. These results underscore KANDI’s potential to address key challenges in offline RL for healthcare applications, offering an effective solution for activity promotion intervention strategies in healthcare.
\end{abstract}


\section{Introduction}
\label{sec:intro}
Physical activity monitoring in free-living environments using wearable sensors, such as accelerometers, provides rich data for understanding human behavior, especially in populations at high health risk.
The advent of wearable sensor technologies has transformed physical activity monitoring by enabling continuous, high-resolution, unobtrusive, and real-time assessment of individual detailed movement patterns ~\cite{Sun2018, hardeman2019systematic, Bezold2021}. Integrating wearable sensor technologies with Artificial Intelligence (AI) is revolutionizing the healthcare landscape. Specifically, Reinforcement Learning (RL), as a subset of AI, plays a pivotal role in enhancing health outcomes through personalized and adaptive interventions~\cite{Qian2022MRT}. 

Fall, a major public health concern, is closely associated with physical inactivity and sedentary behavior, which exacerbate balance issues and increase fall-related injuries, significantly affecting the quality of life among older adults \cite{cdc_falls_2024}.
Research has shown that increasing physical activity can significantly reduce fall risk by improving balance, muscle strength, and overall mobility \cite{sherrington2020evidence}.
Promoting physical activity is a proven strategy for fall prevention, yet implementing effective interventions remains a global challenge. With aging populations growing rapidly, the urgency for accurate and impactful fall prevention strategies has never been greater.
To address these challenges, the \textbf{P}hysio-f\textbf{E}edback \textbf{E}xercise p\textbf{R}ogram (PEER) study has emerged as a promising technology-based body and mind intervention for fall prevention~\cite{Thiamwong2020physio,Thiamwong2023}.
PEER is a two-arm randomized controlled clinical trial integrating visual feedback, cognitive reframing, and peer coaching for fall prevention in older adults. By leveraging wearable technology, it generates rich data ideal for offline RL, a proven approach in AI-driven health applications \cite{tang2022leveraging}.
Building on this foundation, mobile Health (mHealth) interventions, accessible via mobile devices such as wearable sensors, integrate seamlessly into individuals’ daily lives to reduce fall risks by promoting physical activity~\cite{Luers2019, sherrington2020evidence}. 
This fusion of technology and behavioral science paves the way for innovative health decision-making.

Using wearable sensor data from the PEER study, we aim to utilize AI-driven offline RL to learn the optimal timing and policy for encouraging behaviors such as sitting less and moving more~\cite{franklin2010move}. This approach promotes physical activity in older adults and ultimately reduces fall risk.
However, implementing AI in fall prevention is also challenging. 
Firstly, one primary challenge is defining appropriate reward functions that accurately capture the complex, interrelated factors influencing fall risks. In the healthcare setting, rewards are often unobservable and may depend on long-term outcomes, making it difficult to design reward structures that effectively guide AI models toward desired behaviors. 
Secondly, unlike other domains where outcomes are quickly observable, the effects of fall prevention strategies may take a long time to manifest. This delay in feedback makes it difficult for AI models to iteratively refine their policies and adapt to individual participant needs in real time.
Finally, our study observes sensor activity data at high time resolution while incorporating continuous-time actions to capture the dynamics of physical activity. 
In dynamic settings like fall prevention and physical activity promotion, responses fluctuate significantly over time, demanding adaptive models capable of real-time adjustments.

\noindent{\textbf{Related Work}} 
Reinforcement learning has shown considerable promise in healthcare, especially for optimizing intervention policies and promoting physical activity~\cite{Yu2021, Liao2020, Carpenter2020}. In particular, offline RL has revolutionized healthcare applications~\cite{fatemi2022semi, Zhang2023, Brandfonbrener2022}, by enabling intervention optimization using pre-collected data, and minimizing the need for direct environmental interactions~\cite{Yu2021, Lange2012}. 
However, existing methods often struggle with temporal variability, reward inference, and policy optimization~\cite{Hua2022}. While policy optimization methods like Twin Delayed Deep Deterministic Policy Gradient (TD3)~\cite{fujimoto2018addressing}, Advantage-Weighted Actor-Critic (AWAC)~\cite{nair2020awac}, and Implicit Q-learning (IQL)~\cite{kostrikov2021offline} improve sample efficiency, but rely on predefined reward structures. More expressive methods, Diffusion Q-Learning (Diffusion-QL)~\cite{wang2022diffusion} and Efficient Diffusion Policies (EDP)~\cite{Kang2023} improve action modeling but still require well-specified rewards and cannot inherently infer the latent rewards. Although inverse RL (IRL) can enhance pattern predictions to infer the reward function \cite{Arora2021, Grze2017}, it needs refinement for time-varying physical activities across different participants. 

\noindent{\textbf{Proposed Method and Insight}} 
Overcoming the above-mentioned obstacles requires innovative approaches to reward learning, robust methods for handling delayed feedback in policy, and frameworks that account for the indirect nature of healthcare outcomes. 
Our proposed method, \textbf{K}olmogorov-\textbf{A}rnold \textbf{N}etworks and \textbf{D}iffusion Policies for Offline \textbf{I}nverse Reinforcement Learning (KANDI), is designed to achieve two key goals: \textcircled{1} accurately inferring reward functions, and \textcircled{2} capturing continuous-time action distribution and generating high-quality policy trajectories.
To achieve \textcircled{1}, KANDI leverages KAN to infer reward functions from expert (e.g., low-fall-risk participants) behavior, which are particularly adept at handling the complexity of healthcare environments using IRL \cite{Janner2022, Zhou2024}. 
To achieve \textcircled{2}, we integrate the diffusion model into the KANDI framework to address the challenge of distribution shifts, which occur when policies trained on offline data fail to generalize to new scenarios, a common issue in healthcare applications \cite{Bucher2024, Du2024, Janner2022}. Diffusion models mitigate this problem by generating high-fidelity individualized action distributions and producing high-quality policy trajectories \cite{Kang2023, Liu2024, Hao2024}. 

\noindent{\textbf{Our Contributions}} 
Our approach learns the optimal timing and policy to increase physical activity in older adults by promoting moderate-to-vigorous physical activity (MVPA) and reducing sedentary time, with several notable advantages:
\begin{itemize}
    \item We proposed a novel IRL approach using Kolmogorov-Arnold Networks to infer the reward function from expert behavior in free-living healthcare environments where rewards are not explicitly defined. KAN's learnable activation functions effectively handle the complexity and dimensionality of physical activity data, enabling accurate reward inference. 
    \item We use the diffusion model to capture high-dimensional, multi-time-resolution policies in an offline RL healthcare environment. By generating individualized actions from the diffusion policy model instead of directly sampling from the action-state space, we greatly improve learning efficiency while preserving high fidelity in offline RL.
    \item  KANDI maintains high fidelity in real-world clinical trial settings. This innovative approach not only captures the dynamics of free-living environments more accurately but also aligns closely with actual, observable physical activity at specific times. By treating time as a continuous variable, KANDI adapts flexibly to diverse behavioral patterns, including periodic patterns, random wearing durations, and missing data patterns, enhancing its applicability in various clinical scenarios.
\end{itemize}

\section{Problem Formulation}\label{sec:problem}
In this section, we provide background information on the use of wearable sensors in the free-living environment of our \textbf{P}hysio-f\textbf{E}edback \textbf{E}xercise p\textbf{R}ogram (PEER) study, which examines the effectiveness of physical activity promotion and fall prevention among older adults. We then utilize the proposed offline RL framework to identify policies for physical activity engagement.

\begin{figure}[h]
\centering
\centerline{\includegraphics[width = 0.7\linewidth]{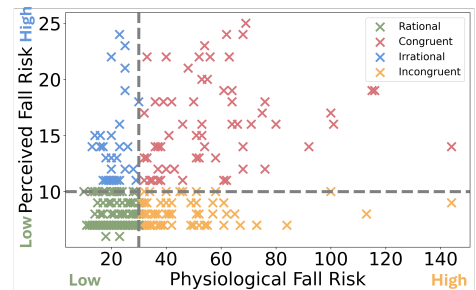}}
\caption{Grouping of Participants into Four Fall Risk Categories Based on Technology-Based Appraisal: The \textit{Rational} Group (Green) Exhibits Low Perceived and Physiological Fall Risk, Serving as Expert Trajectories for Learning.}
\label{fig:tech_feedback}
\end{figure}

\begin{figure*}[t]
    \centering
    \centerline{\includegraphics[width = \linewidth]{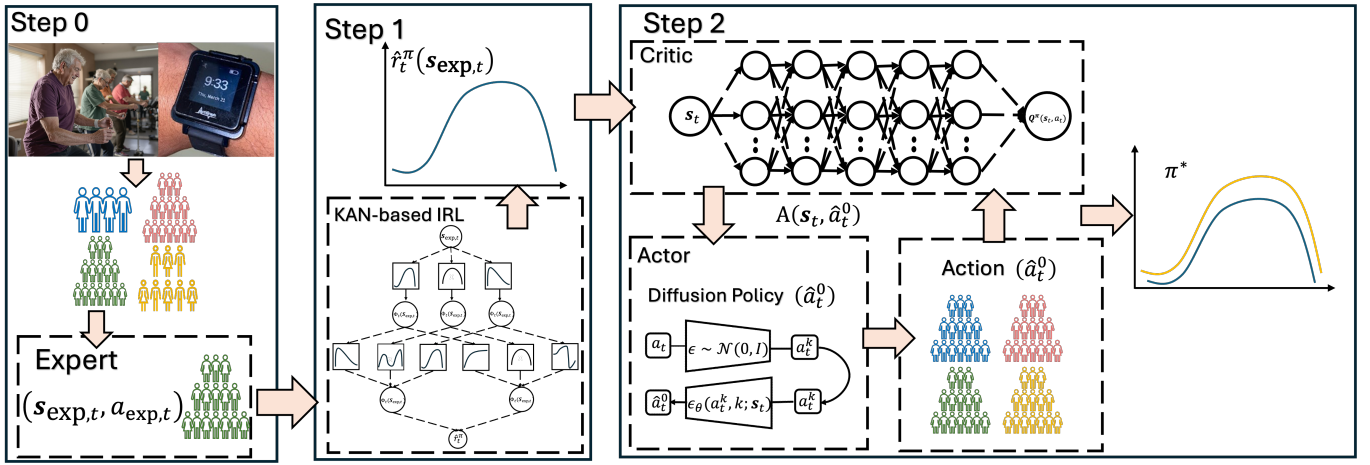}}
    \caption{Overview of the KANDI Framework: Step 0 - Grouping Participants by Fall Risk and Extracting Expert Behavior; Step 1 - Integrating Kolmogorov-Arnold Networks (KAN) into Inverse Reinforcement Learning (IRL) for Reward Inference; Step 2 - Utilizing Diffusion Policies for Action Refinement and Optimizing Policy to Enhance Physical Activity.}
    \label{fig:overview}
\end{figure*}

\subsection{Data Collection Environment Setup}
The PEER physical activity data is derived from a two-arm clustered randomized controlled trial for fall prevention, where participants wear a 3-axis ActiGraph LEAP accelerometer for seven-day cycles to provide minute-level data on activities, steps, and body positions such as lying, sitting, and standing in a free-living environment, both at baseline (T1) and post-30-day intervention (T2). This dataset comprises 2.77 million observations from 134 older adults (aged 61-89 years) across 11 communities in Central Florida, USA, each with diverse fall histories. Participants were categorized into four fall risk categories, namely \textit{Rational}, \textit{Irrational}, \textit{Congruent}, and \textit{Incongruent} (Step 0 of Fig.~\ref{fig:overview}), based on a combination of physiological fall risk (measured using BTracks Balance Test score, poor ability means $\ge 30$) and perceived fall risk (measured using the short Fall Efficacy Scale International (FES-I) score, increased fear of falling means $\ge 10$), as illustrated in Fig.~\ref{fig:tech_feedback} \cite{Thiamwong2023}. 
In the PEER study, participants were randomly assigned to intervention or control groups, creating natural discrepancies since only the intervention group received the program, highlighting heterogeneity in activity trajectories for policy learning.

To preserve the temporal dependence, we organize the physical activity time series using a 30-minute look-back window that includes lagged data alongside the current minute-level physical activity data. We visualize the reward and policy as functions of time over a full 24-hour period for clearer interpretation.
This rich dataset offers opportunities to develop policies that vary with time and to identify the time periods to promote physical activity, ultimately reducing fall risk among older adults while considering individual heritability and fall risk levels. 
Based on these measurements and actions (e.g., standing), our goal is to optimize the policy function based on human behavior patterns and establish a foundation for designing a micro-randomized trial using the learned policy.

\subsection{Objectives}
Our primary objective is to promote physical activity levels among older adults by maximizing cumulative reward learned from experts (i.e., low fall-risk participants).
We use the body position \textit{standing} measured by a gyroscope in the accelerometer as the \textit{action}. The action space $a_t$ was binary, denoting whether the participant was standing ($a_t = 1$) or not ($a_t = 0$).  Standing is strongly associated with reduced sedentary behavior, and we assume that once a participant stands, the probability of engaging in physical activity increases. We define standing status based on body position data aggregated over one-minute intervals. 
Each participant $i\  (i=1,\ldots,n)$ thus has a stream of time-stamped physical activity data triplets, represented as  
\(\{(a_t, \mathbf{s}_t, r_t)\}_{t=1}^T\).  
At each time point \(t\), the action \(a_t\) (i.e., standing \textit{vs}. not) is recorded, while the states \(\mathbf{s}_t\) represent a collection of physical activity metrics, including step counts and vector magnitudes (VMs) from 3-axis accelerometers, and participant characteristics such as enrollment duration, intervention assignment (PEER \textit{vs}. control), and fall risk group (Fig.~\ref{fig:tech_feedback}). The corresponding reward at time \(t\) is denoted as \(r_t\).

The analysis has two main phases: reward learning and policy learning. In Step 1 (Fig.~\ref{fig:overview}), we utilized data from the \textit{Rational} group (low fall-risk participants) as experts to infer the reward function. 
In Step 2 (Fig.~\ref{fig:overview}), KANDI maximizes rewards to determine the optimal probability and timing for standing, thereby enhancing physical activity levels.
Suppose the probability of taking the standing action at time $t$ is modeled by a policy $\pi_t= \pi_t(a | \mathbf{s}): \mathcal{S} \to \mathcal{A}$. Our objective is to learn an optimal policy $\pi^*$ that maximizes the expected cumulative reward over time, $\pi_t^* = \arg \max_{\pi_t} \mathbb{E} \left [ \sum_{\tau} r(\mathbf{s}_\tau, a_\tau)|\pi_t \right]$, where $r(\mathbf{s}_\tau, a_\tau)$ represents the reward function that needs to be learned from Step 1 using IRL, which is associated with the state $\mathbf{s}_\tau$ and action $a_\tau$ at time $\tau$.

\section{\textbf{K}olmogorov-\textbf{A}rnold \textbf{N}etworks and \textbf{D}iffusion Policies for Offline \textbf{I}nverse Reinforcement Learning (KANDI) Framework}\label{sec:kandi}
The KANDI framework (Fig.~\ref{fig:overview}) is designed to address reward inference and policy optimization in healthcare applications. KANDI integrates two key innovations: (1) Kolmogorov-Arnold networks, which model complex, high-dimensional relationships between states, actions, and rewards using learnable activation functions and B-spline mappings for accurate reward inference from expert trajectories; and (2) Diffusion Policies, which sample actions from a diffusion model for robust exploration of the state-action space. By combining within an Actor-Critic framework, KANDI optimizes policies for promoting physical activity (e.g., standing behavior) while generalizing for other decision-making problems, enhancing both the accuracy and efficiency of offline reinforcement learning in dynamic healthcare environments.

\subsection{Kolmogorov-Arnold Networks-based Inverse Reinforcement Learning}\label{KAN}
In healthcare settings, particularly for fall prevention in older adults, defining an explicit reward function directly based on human behaviors in free-living environments presents a significant challenge. 
To address this, we propose utilizing proximal outcomes to measure participants' fall risk (as illustrated in Fig.~\ref{fig:tech_feedback}) and subsequently employ Inverse Reinforcement Learning (IRL) to infer the reward function of an agent based on its observed behavior, such as physical activity levels and participant fall risk \cite{Arora2021}. By leveraging IRL, we can infer the underlying reward structure that governs behaviors like standing, thereby enabling the optimization of the policy $\pi_t^*$ to effectively mitigate fall risk.
By formulating the IRL problem, we use observational behavior to infer the hidden preference, thus avoiding the need for manual specification of the reward function \cite{Ng2000}. This helps mitigate biases arising from misspecified rewards. Since the exact policy $\pi_t$ is unknown, we access it by learning from a set of actual trajectories within the state space. Specifically, we treat the \textit{Rational} group, which is a subset of individuals with minimal fall risk at baseline in the PEER study, as the expert trajectories.


%

We used the Kolmogorov-Arnold Networks instead of multi-layer perceptrons to model the complex relationship between states and actions, and learn the reward function in IRL. This approach overcomes the limitations of conventional methods in capturing these intricate relationships. KAN is based on the Kolmogorov-Arnold representation theorem, which approximates multivariate continuous functions using a sum of simpler, learnable functions \cite{Liu2024}. 

The reward function $r_t$ at any state observed from the expert trajectory $\mathbf{s}_{\text{exp},t}$ is estimated using a weighted sum of transformed state variables, where transformations are governed by both the activation functions and the B-spline mappings. This allows us to integrate and adapt features from the expert state space efficiently, which is crucial for accurate reward prediction, so we enhance the estimated value function $\hat{r}^{\pi}_t(\mathbf{s}_{\text{exp},t})$ as follows, 
\begin{equation}\label{eq:kan}
  \hat{r}^{\pi}_t(\mathbf{s}_{\text{exp},t}) = \sum_{t=\tau}^T \alpha_t \gamma^{t-1} \sum_{l=1}^k w_l \sigma\left(\sum_{j=1}^m v_{lj} B_j(\mathbf{s}_{\text{exp},t})\right)
\end{equation} 
where $\sigma(\cdot)$ is the Sigmoid Linear Unit (SiLU) activation function, $w_l$ represents weights for the $l$-th neurons in the output layer, $v_{lj}$ is weights connecting the $j$-th input feature to the $l$-th neuron in the hidden layer, $k$ is the number of neurons in the hidden layer, and $m$ is the number of input features of the state vector. $B_j$ are B-spline functions that map the expert states, allowing multiple state variables, $\mathbf{s}_{\text{exp},t} = \{s_{\text{exp},t}^1, s_{\text{exp},t}^2, \cdots, s_{\text{exp},t}^m\}$, including times (in minutes), body positions, lagged physical activity time series and steps, PEER intervention group assignments, fall risk appraisal groups, to be considered simultaneously. To effectively approximate continuous reward functions, we incorporate the B-spine basis function with several knots, balancing computational efficiency and flexibility in capturing complex reward structures~\cite{Liu2024}.
Since physical activity states vary over time, we introduce a time-varying parameter $\alpha_t$ to account for the differences between daytime and nighttime. Specifically, $\alpha_t$ is positive during the daytime and negative during the nighttime.
The formulation Eq.~\eqref{eq:kan} focuses on the weighted contributions of transformed expert state features at each time step, effectively streamlining the representation.

The utilization of KAN improves the accuracy of reward function representations, thereby enhancing our ability to learn from the expert behaviors of the \textit{Rational} group, a targeted low fall risk group in the PEER study. 
In this context, the primary objective of using KAN is to accurately estimate and optimize the reward function based on observed data from expert agents. The KAN-based IRL framework trains the network to minimize the difference between the predicted rewards and the expert behavior, typically using a maximum entropy framework to naturally capture variability and stochasticity in expert behavior, as illustrated in Algorithm~\ref{al:kan}. This approach aligns the learned reward function with expert behavior by incorporating entropy regularization,
\begin{equation}
    \text{max}\  \mathbb{E}_{\pi_\theta} \left[ \sum_{t=1}^T \gamma^{t-1} \Phi_t (\mathbf{s}_{\text{exp},t}) + \beta H(\pi_\theta(a_{\text{exp},t} | \mathbf{s}_{\text{exp},t}))\right],
\end{equation}
where $H(\pi_\theta(a_{\text{exp},t} | \mathbf{s}_{\text{exp},t})) \\
=  - \sum_{a_{\text{exp},t}} \pi_\theta (a_{\text{exp},t}|\mathbf{s}_{\text{exp},t}) \log \pi_\theta (a_{\text{exp},t}|\mathbf{s}_{\text{exp},t})$,  represents the entropy of the policy, and $\beta$ is a regularization parameter that control the trade-off between matching and the maximizing entropy.

\begin{algorithm}
\caption{KAN-Based IRL for Reward Functions Inference}\label{al:kan}
\begin{algorithmic}[1]
\Require Expert behavior $\mathbf{s}_{\text{exp},t}$, discount factor $\gamma$, time-vary parameter $\alpha_t$, regularization parameter $\beta$, learning rate $\eta$
\For{each participants $i$}
    \For{each time step $t$ in $\tau_i$}
        \State $\Phi_t(s_{\text{exp},t}) = \sum_{l=1}^{k} w_l         
            \sigma\left(\sum_{j=1}^{m} v_{lj} B_j(s_{\text{exp},t})\right)$
        \State $\hat{r}^{\pi}_t(\mathbf{s}_{\text{exp},t}) = \sum_{t=\tau}^{T} \alpha_t \gamma^{t-1} \Phi_t(s_{\text{exp},t})$
        \State $\Phi \gets \Phi + \eta \nabla_{\Phi} \mathbb{E}_{\pi_{\theta}} \left[ \sum_{t=1}^{T} \gamma^{t-1} \Phi_t(\mathbf{s}_{\text{exp},t}) - \beta H(\pi_{\theta}) \right]$
    \EndFor
\EndFor
\State \Return Reward function $\hat{r}^{\pi}_t(\mathbf{s}_{\text{exp},t})$
\end{algorithmic}
\end{algorithm}

\subsection{Diffusion Policies}\label{DP}
We integrate the diffusion model into the Actor-Critic RL framework to learn the optimal policies for enhancing physical activity. 
The diffusion model generates high-fidelity actions by progressively adding and removing noise, enabling stable exploration and enhancing policy performance while efficiently refining actions.
Instead of sampling action directly from the policy distribution, the diffusion policy generates individual actions that closely align with the distribution of observed behavior, thereby improving both accuracy and efficiency during training. 
In our diffusion Actor-Critic framework (Step 2 of Fig.~\ref{fig:overview}), the actor network uses the diffusion policy to generate actions, and these actions are then refined and evaluated by the critic network. Particularly, the diffusion policy follows a two-step process, (1) the forward step that iteratively adds Gaussian noise to actions initially sampled from the actor network, so that we can avoid the abrupt changes and improve overall system stability, and (2) the reverse step that removes Gaussian noise and refine actions to closely align with the policy to be optimized~\cite{ho2020denoising, Kang2023}. 

In the forward process of diffusion policy, initial actions $a_t^0$ are firstly sampled from the actor network, which selects the best action from the current state. To ensure smooth policy learning and foster exploration, these actions are iteratively perturbed with Gaussian noise. At any $n$-th step, the noisy action $a_t^n$ follows Gaussian distribution, $\mathcal{N} (a_t^n; \sqrt{\bar{\eta}^n} a_t^0, (1-\Bar{\eta}^n)\boldsymbol{I})$, where $\bar{\eta}^n$ is noise scaling factor that gradually decreases over $N$ diffusion steps to control noise decay. This controlled noise decay enables more stable action exploration, maintaining the overall structure of the policy. Consequently, we compute the perturbed actions as $a_t^n = \sqrt{\Bar{\eta}^n} a_t^0 + \sqrt{1- \Bar{\eta}^n} \epsilon,~\epsilon \sim \mathcal{N} (0, \boldsymbol{I})$.

The reverse process of the diffusion policy involves learning a noise schedule and gradually removing Gaussian noise based on this learned schedule. This allows us to generate refined actions that closely align with the learned policy.
Instead of directly reversing the noise, we used the actor network to predict the optimal noise term $\epsilon_{\theta} (a_t^n, n; \mathbf{s}_t)$ based on the current state $\mathbf{s}_t$ and the action from forward process $a_t^n$. The predicted noise helps reconstruct the approximate original action to improve policy performance,
\begin{equation}
    \hat{a}_t^0 =(\bar{\eta}^n)^{-1/2} \left(a_t^n - \sqrt{1 - \bar{\eta}^n} \epsilon_{\theta}(a_t^n, n; \mathbf{s}_t)\right).
\end{equation}
To train the optimal noise term $\epsilon_{\theta} (a_t^n, n; \mathbf{s}_t)$, the actor network optimizes a denoising loss that directly compares the true action $a_t$ and the approximated action $\hat{a}_t^0$, equivalent to compare the noise term, $\epsilon$, and predicted noise term, $\epsilon_{\theta} (a_t^n, n; \mathbf{s}_t)$. This approach improves training efficiency by sidestepping the need to explicitly model the original noise. Consequently, the goal of the actor network is to minimize the following objective, $\ell(a_t) = \mathbb{E}\Bigl[\| a_t - \hat{a}_t^0 \|^2 \Bigr]$, where $\hat{a}_t^0$ is obtained from the noisy action $a_t^n$ using predicted noise $\epsilon_{\theta}$. 

We then use the critic network to evaluate the actions $\hat{a}_t^0$ generated by the actor network after establishing the diffusion policy in the actor network.
The critic network uses action-value function $Q^{\pi} (\mathbf{s}_t, \hat{a}_t^0) = \hat{r}^{\pi}_t(\mathbf{s}_t) + \gamma V^{\pi}(\mathbf{s}_{t+1})$, where $\hat{r}^{\pi}_t(\mathbf{s}_t)$ is the inferred rewards from KAN-based IRL, and $V^{\pi}(\mathbf{s}_{t+1})$ is the estimated state value function of the next state $\mathbf{s}_{t+1}$ under the current policy $\pi$~\cite{Cao2024}. The critic network ensures the final selected action aligns with the reward patterns, improving policy performance. 

The goal of the diffusion policy within the Actor-Critic framework is to optimize the actor network by maximizing the expected rewards while stabilizing action selection through a denoising process. In general, policy optimization is performed by weighted maximum likelihood estimation. In diffusion policy, we can simplify the objective by the approximated action $\hat{a}_t^0$, enhancing the stability and efficiency of the diffusion process, 
\begin{equation}
    \underset{\pi_\theta}{\text{max}} \ \mathbb{E} \left[ A(\mathbf{s}_t, \hat{a}t^0) \log \pi{\theta} (\hat{a}_t^0 | \mathbf{s}_t) \right] - \lambda \mathbb{E} \left[ \| a_t - \hat{a}_t^0\|^2 \right], 
\end{equation}
where $A(\mathbf{s}_t, \hat{a}^0_t) = Q^{\pi}(\mathbf{s}_t, a_t) - V^{\pi}(\mathbf{s}_t)$ is advantage function, where $V^{\pi}(\mathbf{s}_t) = \mathbb{E}_{\pi_{\theta}} \left[ \sum_{n=0}^{T} \gamma^n \hat{r}^{\pi}_{t+n} | \mathbf{s}_t\right]$, estimating the exacted return of being in state $\mathbf{s}_t$ under the current policy. And $\log \pi_{\theta} (\hat{a}^0_t | \mathbf{s}_t)$ is the log probability of approximated action $\hat{a}^0_t$ under policy $\pi_\theta$, ensuring that better actions are more likely to be chosen. The $\lambda$ is a hyperparameter controlling the balance between policy optimization and action refinement. 

\begin{algorithm}
\caption{Diffusion Policy in Actor-Critic RL}
\begin{algorithmic}[1]
\Require All participants $\{(a_t, \mathbf{s}_t, \hat{r}^{\pi}_t(\mathbf{s}_t))\}_{t=1}^T$, 
Regularization $\lambda$, Learning rate $\eta_{\theta}$, Number of diffusion steps $N$, Discount rate $\gamma$
\For{$t = 1, \dots, T$}
    \State Initial action $a_t^0 \sim \pi_{\theta}(a_t | \mathbf{s}_t)$
    \For{diffusion step $n = 1, \dots, N$} \Comment{Diffusion Policy}
    \State $a_t^n = \sqrt{\bar{\eta}^n} a_t^0 + \sqrt{1 - \bar{\eta}^n} \epsilon, \quad \epsilon \sim \mathcal{N}(0, I)$
\EndFor
\For{$n=N, \dots, 1$}
    \State $\hat{a}_t^0 =(\bar{\eta}^n)^{-1/2} \left(a_t^n - \sqrt{1 - \bar{\eta}^n} \epsilon_{\theta}(a_t^n, n; \mathbf{s}_t)\right)$ \Comment{$\epsilon_{\theta}(a_t^n, n; \mathbf{s}_t)$ trained from the actor network}
\EndFor
    \State $V^\pi(\mathbf{s}_t) = \mathbb{E}_{\pi_{\theta}} \left[ \sum_{n=0}^{T} \gamma^n \hat{r}^{\pi}_{t+n} | \mathbf{s}_t \right]$ \Comment{Critic Network Update}
    \State $Q^{\pi}(\mathbf{s}_t, \hat{a}_t^0) = \hat{r}^{\pi}_t(\mathbf{s}_t) + \gamma V^{\pi}(\mathbf{s}_{t+1})$
    \State $A(\mathbf{s}_t, \hat{a}_t^0) = Q^{\pi}(\mathbf{s}_t, \hat{a}_t^0) - V^{\pi}(\mathbf{s}_t)$
    \State $\theta \gets \theta + \eta_\theta \nabla_\theta \mathbb{E}_{\pi_{\theta}} \left[ A(\mathbf{s}_t, \hat{a}_t^0) \log \pi{\theta} (\hat{a}_t^0 | \mathbf{s}_t) \right] - \lambda \ell(a_t)
    $ 
\EndFor
\State \Return Policy $\pi_{\theta}(a_t | \mathbf{s}_t)$
\end{algorithmic}
\end{algorithm}

\section{In-field Experiments on Physio-fEedback Exercise pRogram (PEER) study}\label{sec:peer}
\subsection{Data \& Environment Setup}
We implement the proposed KANDI framework to learn policies for promoting physical activity in the PEER study (see details in Section~\ref{sec:problem}), providing valuable insights into free-living activity patterns, serving as a solid foundation for evaluating the effectiveness of our fall prevention strategy. 

Our objective is to learn a realistic reward function from the target population experience, and to determine how effectively the trained policies can promote physical activity and identify the optimal time for implementing fall prevention interventions among older adults with varying levels of fall risk.
The KAN architecture in the KANDI framework consists of a three-layer neural network for reward inference in IRL, using the SiLU activation function, Adam optimizer, and B-spline basis functions with 10 knots,
allowing KAN to effectively capture the underlying reward structures in high-dimensional state spaces. The Diffusion Policy employs the action refinement mechanism with 100 diffusion steps and linear decay to stabilize action selection, optimized with Adam (Weight Decay, learning rate 0.0001) and batch size 256. 
Our training environment operates on a cluster with CUDA acceleration, utilizing three NVIDIA RTX A2000 12GB GPUs, with CUDA version 11.3 and Python 3.8.10. 

\begin{figure}[t]
    \centering
    \centerline{\includegraphics[width = 0.8\linewidth]{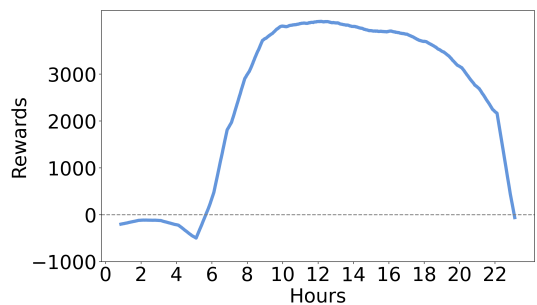}}
    \caption{Temporal Reward Function at Minute Level Learned Using KAN-Based Inverse Reinforcement Learning with the \textit{Rational} Group as Expert Trajectories (Step 1 of Figure~\ref{fig:overview}).}
    \label{fig:reward}
\end{figure}

\subsection{Results: Reward Function Learning}
We trained the KAN-based IRL model using the \textit{Rational} subpopulation group with low fall risk, which included data from 58 participants and comprised 1.07 million observations at minute-level intervals over 7 days for both the pre- and post-PEER intervention periods.

Since our ultimate objective was to enhance physical activity levels based on daily activity patterns, we focused on optimizing rewards during the daytime while reducing them at nighttime. Specifically, when a participant's physical activity level reached Moderate to Vigorous Physical Activity (MVPA) status, defined by VMs exceeding $1100$ per minute during daylight hours, we considered the rewards as positive. Conversely, during nighttime when physical activity levels typically dropped to sedentary (VMs $< 600$), rewards were negatively adjusted to reflect lower activity expectations. 
The rewards function plot, as shown in Figure \ref{fig:reward}, demonstrated the model's effective captures and representation of activity patterns. Throughout the day, from $6$ a.m. to $10$ p.m., rewards remained positive, peaking between $8$ a.m. to $6$ p.m., and began to taper off post-$6$ p.m. By effectively deriving the reward function, it laid a solid foundation for policy learning phase aimed at promoting physical activity levels. 


\begin{figure}[t]
\centering
\centerline{\includegraphics[width = 0.7\linewidth]{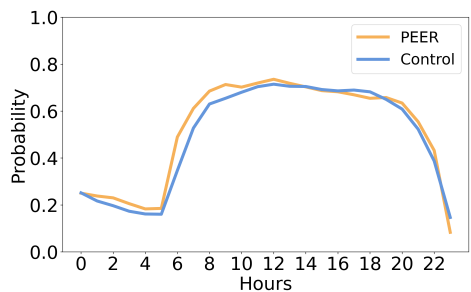}}
\caption{Policy of Standing Action Over Time (Minute Level) for PEER vs. Control Groups, Learned Using Diffusion Policies (Step 2 of Figure~\ref{fig:overview}) Based on Learned Rewards.} 
\label{fig:action by hour}
\end{figure}

\begin{figure}[th]
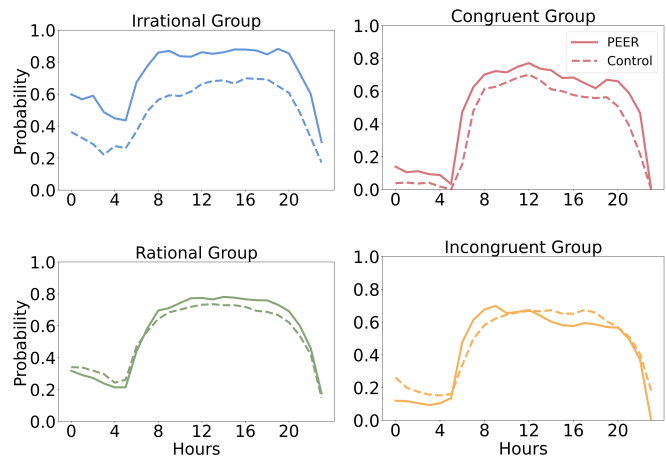

\centering
\subfloat{\includegraphics[width=0.49\linewidth]{irrational_prob_over_hours_color.pdf} 
    \label{subfig:irrational}} \hfill    \subfloat{\includegraphics[width=0.475\linewidth]{congruent_prob_over_hours_color.pdf}
    \label{subfig:congruent}} \\
\subfloat{\includegraphics[width=0.49\linewidth]{rational_prob_over_hours_color.pdf}
    \label{subfig:rational}}
    \hfill    \subfloat{\includegraphics[width=0.475\linewidth]{incongruent_prob_over_hours_color.pdf}
    \label{subfig:incongruent}}
\caption{Policy of Standing Action Over Time (Minute Level) for Four Fall Risk Groups, Highlighting How KANDI Reinforces Action Probabilities of Other Groups Toward the \textit{Rational} Group (Expert).}
\label{fig:action by fall risk group}
\end{figure}

\begin{table*}
\centering
\begin{threeparttable}
\caption{Average normalized score on the D4RL benchmark under predefined reward functions and latent dynamic rewards.}
\label{table1}
\scriptsize
\begin{tabular}{c|ccccccc|cccc}
\toprule
\textbf{Dataset} & \multicolumn{7}{c|}{Predefined Reward Functions} & \multicolumn{4}{c}{Latent Dynamic Reward Functions with KAN-based IRL} \\
\cmidrule(lr){2-8} \cmidrule(lr){9-12} &
\textbf{BC} &
\textbf{TD3} & 
\textbf{AWAC} & 
\textbf{IQL} & 
\textbf{Diffusion-QL} & 
\textbf{EDP} & 
\textbf{KANDI} &
\textbf{KAN+TD3} & 
\textbf{KAN+AWAC} & 
\textbf{KAN+IQL} & 
\textbf{KANDI} \\
\midrule
halfcheetah-medium-v2 & 42.6 & 48.4 & 43.5 & 47.4 & 51.5 & \textbf{52.8} & 51.8 
& 49.5 & 47.1 & 50.2 & \textbf{52.6} \\
walker2d-medium-expert-v2 & 107.5 & 90.7 & 74.5 & 109.6 & 111.2 & \textbf{112.0} & 111.4 
& 92.3 & 76.5 & 110.1 & \textbf{111.8} \\
hopper-medium-expert-v2 & 52.5 & 98.0 & 55.8 & 91.5 & \textbf{112.3} & 112.0 & 112.1 
& 95.2 & 57.3 & 92.0 & \textbf{112.0} \\
\midrule
kitchen-complete-v0 & 64.8 & 2.2 & 39.3 & 62.5 & 84.5 & 97.0 & \textbf{97.3} 
& 5.8 & 41.2 & 63.0 & \textbf{96.4} \\
kitchen-mixed-v0 & 50.5 & 0.0 & 22.0 & 51.0 & 66.6 & 73.0 & \textbf{74.0} 
& 1.2 & 26.7 & 53.4 & \textbf{74.3} \\
\midrule
antmaze-large-play-v0 & 0.0 & 0.2 & 0.0 & 39.6 & 49.0 & 60.0 & \textbf{62.5} 
& 0.3 & 1.0 & 42.0 & \textbf{66.1} \\
antmaze-large-diverse-v0 & 0.0 & 0.0 & 1.0 & 47.5 & 61.7 & \textbf{64.0} & 63.7 
& 0.1 & 3.5 & 49.2 & \textbf{63.9} \\
\bottomrule
\end{tabular}%
\begin{tablenotes}
\footnotesize
\item *BC results are taken directly from EDP \cite{Kang2023}. Other baselines are re-implemented. 
KANDI achieves competitive or superior performance across both predefined and latent dynamic reward settings, particularly in the Kitchen and AntMaze domains.
\end{tablenotes}
\end{threeparttable}
\end{table*}

\subsection{Result: Diffusion Policies Learning}
We focus on the policy training phase (Step 2 of Fig.~\ref{fig:overview}), where we analyze the policy learned as a function of time, intervention group assignment (PEER \textit{vs}. control), and fall risk categories among a total of 134 participants.
The policy reveals that the probability of standing is very low during the early morning and late night, peaking around noon and remaining high until sunset in Fig.~\ref{fig:action by hour}.
This pattern aligns with the daily activity rhythms of older adults ~\cite{xie2023smoothing}. The PEER group also shows higher action probabilities than the control group, especially during daytime, highlighting the intervention’s efficacy in promoting physical activity~\cite{liu2025effectiveness}.

If we break down the policy by fall risk categories, as shown in Fig. \ref{fig:action by fall risk group}, we can observe more detailed temporal patterns and differences between the PEER and control groups. The \textit{Irrational} group showed a marked increase in activity for the PEER group, indicating that the intervention effectively reduced their high fear of falling.
The \textit{Rational} group, as the expert, shows similar probabilities of action across both PEER and control groups, suggesting that participants with better balance and lower fear of falling are less reliant on the intervention. 
The \textit{Congruent} group benefited from PEER shows higher activity levels in the PEER group, 
reflecting the intervention’s success in engaging those with poor balance and high fear of falling.
However, the \textit{Incongruent} group saw no significant difference between the PEER and control groups, likely due to their poor balance not being immediately improvable through external exercises.

\begin{figure}
\centering
\centerline{\includegraphics[width = 0.7\linewidth]{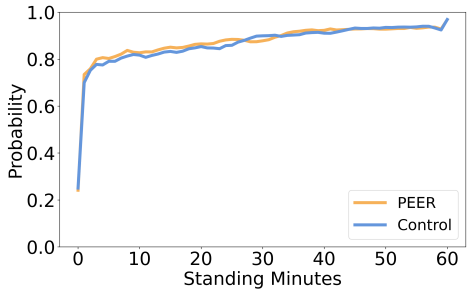}}
\caption{Policy on Standing Minutes per Hour, Demonstrating That Increased Standing Time Can Effectively Enhance Physical Activity Levels. }
\label{fig:action by standing minutes}
\end{figure}

Additionally, we zoom into the trend in action probability based on standing minutes per hour. The action probability increased when standing over 3 minutes in Fig.~\ref{fig:action by standing minutes}. 
Notably, the action probability for the PEER group consistently exceeds that of the control group in the first 30 minutes, underscoring the intervention’s role in promoting longer standing periods, which correlate with enhanced physical activity. 

\section{Experiments on Offline RL Benchmarks}
\noindent\textbf{Baselines and Evaluation Metric}  We conduct evaluations on three domains in the D4RL benchmark~\cite{fu2020d4rl}: Gym locomotion, Kitchen, and Ant Maze. For each domain, we compare a diverse set of strong offline RL baselines to provide a thorough evaluation, including Behavior Cloning (BC)~\cite{bain1995framework}, which is a commonly used baseline that learns a policy on expert actions, TD3~\cite{fujimoto2018addressing}, AWAC~\cite{nair2020awac}, IQL~\cite{kostrikov2021offline}, Diffusion-QL~\cite{wang2022diffusion}, and EDP~\cite{Kang2023}. We use average normalized score, a standard metric for fair cross-environment comparisons for assessment~\cite{fu2020d4rl}. 

\noindent{\textbf{Environment Setup}} 
We adopt a unified network architecture across all tasks for fair comparison. The policy and Q-value networks share a 3-layer MLP backbone with hidden dimension 256 and Mish activation following diffusion-QL~\cite{wang2022diffusion}. For the diffusion policy, the noise prediction network $\epsilon_{\theta}(\mathbf{a}^n, n; \mathbf{s})$ is conditioned on timestep $n$ and concatenated with noisy action $\mathbf{a}^n$ and state $\mathbf{s}$~\cite{Kang2023}. We used Adam to optimize diffusion policy and Q networks. For the KANDI framework, the diffusion policy uses the same noise network, and the reward module is pretrained for 100 epochs and updated during training. Training runs for 2000 epochs on Gym locomotion and 1000 epochs on Kitchen and Ant Maze, with batch size 256 and 15 diffusion steps. Baselines are re-implemented with the same backbone and optimization. Results are averaged over 5 replicates.

\noindent\textbf{Experiment Results} We evaluate the performance of standard policy optimization methods under predefined reward functions and latent dynamic reward functions with KAN-based IRL in Table~\ref{table1}. With predefined reward functions, Ant Maze remains difficult with near-zero baseline results due to sparse expert data. Incorporating KAN-based IRL improves performance in all domains, with clear gains in sparse-reward settings, demonstrating KAN’s effectiveness in reward inference. We partially assessed the role of KAN (Table~\ref{table1}), though a full ablation will be explored in future work. These results highlight that KAN-based IRL effectively recovers reward structures from offline data and that KANDI can robustly generalize to complex, long-horizon tasks.

\section{Conclusion and Future Work}
Our proposed KANDI applied Kolmogorov-Arnold Networks and Diffusion Policies for offline inverse Reinforcement Learning to real-world clinical trial data to optimize the reward function and learn the optimal behavior policy, demonstrating its effectiveness in the AI for the health domain. Our results show that KANDI learned reward functions closely aligned with expert behavior, and that the diffusion-based policy learning captures the multidimensional dynamics and accurately reflects behavioral changes in real-world settings. 

The optimal policy and intervention timing learned from KANDI can inform mobile health solutions, such as micro-randomized trials~\cite{necamp2020assessing} or Just-in-Time Adaptive Interventions (JITAI,~\cite{hardeman2019systematic}), in smart health applications. Future work will extend to larger and more diverse populations and explore integration with multimodal data to enhance generalizability. Overall, KANDI provides a robust foundation for advancing AI-driven health interventions that improve physical activity and intervention efficacy among older adults, with strong potential for adaptation to other healthcare decision-making problems.



\bibliographystyle{ieeetr}






\end{document}